\documentclass[letterpaper]{article} 
\usepackage{aaai25}  
\usepackage{times}  
\usepackage{helvet}  
\usepackage{courier}  
\usepackage[hyphens]{url}  
\usepackage{graphicx} 
\usepackage{natbib}  
\usepackage{caption} 
\usepackage{algorithm}
\usepackage{algorithmic}

\usepackage{amsmath}
\usepackage{amssymb}
\usepackage{amsfonts}
\usepackage{bm}
\usepackage{multirow}
\usepackage{multicol}
\usepackage{graphicx}
\usepackage{subfig}
\usepackage{caption}
\usepackage{xcolor}
\usepackage{rotating}
\usepackage{diagbox}
\usepackage{booktabs}
\usepackage{bbding}

\def\method{C$^3$GNN}

\usepackage{newfloat}
\usepackage{listings}
\DeclareCaptionStyle{ruled}{labelfont=normalfont,labelsep=colon,strut=off} 
\floatstyle{ruled}
\newfloat{listing}{tb}{lst}{}
\floatname{listing}{Listing}
\title{Cluster-guided Contrastive Class-imbalanced Graph Classification}
\author{
    Wei Ju\textsuperscript{\rm 1},
    Zhengyang Mao\textsuperscript{\rm 2},
    Siyu Yi\textsuperscript{\rm 3}\thanks{Corresponding authors.},
    Yifang Qin\textsuperscript{\rm 2},
    Yiyang Gu\textsuperscript{\rm 2},
    Zhiping Xiao\textsuperscript{\rm 4},\\
    Jianhao Shen\textsuperscript{\rm 5},
    Ziyue Qiao\textsuperscript{\rm 6},
    Ming Zhang\textsuperscript{\rm 2}\textsuperscript{*},
}
\affiliations{
    \textsuperscript{\rm 1} College of Computer Science, Sichuan University, Chengdu, China\\
    \textsuperscript{\rm 2} School of Computer Science, State Key Laboratory for Multimedia Information Processing,\\ PKU-Anker LLM Lab, Peking University, Beijing, China\\
    \textsuperscript{\rm 3} College of Mathematics, Sichuan University, Chengdu, China\\
    \textsuperscript{\rm 4} Paul G. Allen School of Computer Science and Engineering, University of Washington, Seattle, WA, USA\\
    \textsuperscript{\rm 5} Huawei Hisilicon, Shanghai, China\\
    \textsuperscript{\rm 6}School of Computing and Information Technology, Great Bay University, Dongguan, China\\
    \{juwei, siyuyi\}@scu.edu.cn, zhengyang.mao@stu.pku.edu.cn, \{qinyifang, yiyanggu, mzhang\_cs\}@pku.edu.cn,\\
    patxiao@uw.edu, jhshen95@gmail.com, zyqiao@gbu.edu.cn
}

\usepackage{bibentry}

\begin{document}

\maketitle

\begin{abstract}
This paper studies the problem of class-imbalanced graph classification, which aims at effectively classifying the graph categories in scenarios with imbalanced class distributions. While graph neural networks (GNNs) have achieved remarkable success, their modeling ability on imbalanced graph-structured data remains suboptimal, which typically leads to predictions biased towards the majority classes. On the other hand, existing class-imbalanced learning methods in vision may overlook the rich graph semantic substructures of the majority classes and excessively emphasize learning from the minority classes. To address these challenges, we propose a simple yet powerful approach called C$^3$GNN that integrates the idea of clustering into contrastive learning to enhance class-imbalanced graph classification. Technically, C$^3$GNN clusters graphs from each majority class into multiple subclasses, with sizes comparable to the minority class, mitigating class imbalance. It also employs the Mixup technique to generate synthetic samples, enriching the semantic diversity of each subclass. Furthermore, supervised contrastive learning is used to hierarchically learn effective graph representations, enabling the model to thoroughly explore semantic substructures in majority classes while avoiding excessive focus on minority classes. Extensive experiments on real-world graph benchmark datasets verify the superior performance of our proposed method against competitive baselines.
\end{abstract}


\section{Introduction}

Graphs are widely recognized as highly effective structured data for representing complex relationships among objects in various domains~\cite{ju2024comprehensive}, including social analysis, bioinformatics, and recommender systems. As such, there is considerable interest in exploring the potential of analyzing graph data, covering a broad spectrum of graph learning tasks, such as node classification~\cite{luo2023toward}, graph classification~\cite{mao2023rahnet}, and graph clustering~\cite{yi2023redundancy}. Among these, graph classification is one of the most interesting also popular topics, which aims to classify the class labels of graphs and has emerged as a significant research focus in various scenarios like molecular property prediction and protein functional analysis.

To well solve this task, graph neural networks (GNNs) \cite{kipf2017semi} have recently emerged as powerful approaches attributed to the message propagation and feature transformation mechanisms~\cite{gilmer2017neural}. However, the success of existing GNN methods typically relies on the assumption that the class distribution in the dataset is balanced. Unfortunately, this assumption fails in many real-world scenarios, where a large portion of classes have a small number of labeled graphs (minority classes) while a few classes have a significantly large number of labeled graphs (majority classes), exhibiting severely skewed class distributions. For instance, the NCI dataset comprises graphs of chemical compounds~\cite{wale2008comparison}, where only approximately 5$\%$ of molecules are labeled as active in the anti-cancer bioassay test, while the remaining molecules are labeled as inactive. This class-imbalanced issue can lead to the notorious prediction bias phenomenon~\cite{zhou2020bbn}, where the GNN classifier favors the majority classes and ignore the minority classes. Consequently, directly applying GNNs to class-imbalanced graphs poses a significant challenge in real-world situations.

Actually, the issue of imbalance in visions has received increasing attention and efforts have been devoted to addressing it over a significant period. In general, the solutions can be categorized into three main groups: re-sampling, re-weighting, and ensembling learning. Re-sampling strategies~\cite{chawla2002smote,han2005borderline,guo2021long} attempt to balance the class distribution by generating synthetic training data. Re-weighting strategies~\cite{cui2019class,cao2019learning,hou2023subclass} focus on adjusting the loss function to assign different weights to the training samples from various classes. For ensembling learning approaches~\cite{xiang2020learning,wang2020long}, they integrate multiple classifiers within a multi-expert framework to achieve robust predictions.

However, directly applying existing class-imbalanced learning methods from vision to graph domains is challenging due to two key limitations. \emph{First}, due to the complexity and diversity of the graph topology, there typically exist hierarchical substructures in graph samples within the majority classes. For example, considering the polarity of molecules, there are numerous naturally occurring molecules with the functional group ``-OH". These molecules possess polarity (class label), but they are further classified into different hierarchical levels based on the strength of their polarity (subclasses). When connected to the functional group ``-C=O", they exhibit strong polarity, whereas when connected to the ``-CH$_2$", they exhibit relatively weak polarity. As a result, within the same class label, the samples of majority class display varying levels of semantic substructure. \emph{Second}, existing re-sampling and re-weighting strategies excessively focus on minority classes, usually at the cost of sacrificing the accuracy of majority classes. However, the samples of minority classes in the graph domain, such as those containing the adamantyl group, are scarce and mostly lack polarity, resulting in a relatively simple semantic structure. These molecules often cannot represent all the semantic structures of the minority classes. Overlearning from such samples introduces certain redundant information and causes prediction bias. \emph{In addition}, although there are many class-imbalanced methods designed for graphs~\cite{shi2020multi,liu2021tail,park2021graphens,zhou2023graphsr,zeng2023imgcl}, they are primarily developed for node-level classification on a single graph. However, the promising yet challenging task of graph-level classification has largely remained unexplored, and it serves as the main focus of this paper.

To address these challenges, in this paper we propose a \underline{C}luster-guided \underline{C}ontrastive \underline{C}lass-imbalanced framework called \method{} for graph classification. The key idea of \method{} is to learn effective graph-level representations by incorporating the principle of clustering into supervised contrastive learning~\cite{khosla2020supervised}. Specifically, to capture hierarchical semantic substructures of the majority classes, \method{} first adaptively clusters graphs of each majority class into multiple subclasses, ensuring that the sample sizes in each subclass are comparable to the minority class. Then, we utilize the Mixup technique to generate synthetic samples, further enriching the semantic information within each subclass and preventing representation collapse due to sparse subclass samples. Based upon this, we leverage supervised contrastive learning to hierarchically learn graph-level representations, encouraging a graph to be (i) closer to graphs from the same subclass than to any other graph, and (ii) closer to graphs from different subclasses but the same class than to graphs from any other subclass. In this way, we are able to capture rich graph semantic substructures of the majority classes and alleviate overlearning in the minority classes, achieving well class-imbalanced learning. To summarize, the main contributions of our works are as follows:

\begin{itemize}
    \item We explore an intriguing and relatively unexplored challenge: class-imbalanced graph classification, with the aim of providing valuable insights for future research.
    \item We propose a novel class-imbalanced learning approach on graphs via clustering and contrastive learning to preserve the hierarchical class substructures as well as achieve balanced learning for all classes.
    \item We conduct extensive experiments to demonstrate that \method{} can provide superior performance over baseline methods in multiple real-world benchmark datasets.
\end{itemize}

\section{Problem Definition \& Preliminaries}

\begin{figure*}[t!]
    \centering
    \includegraphics[width=0.878\textwidth]{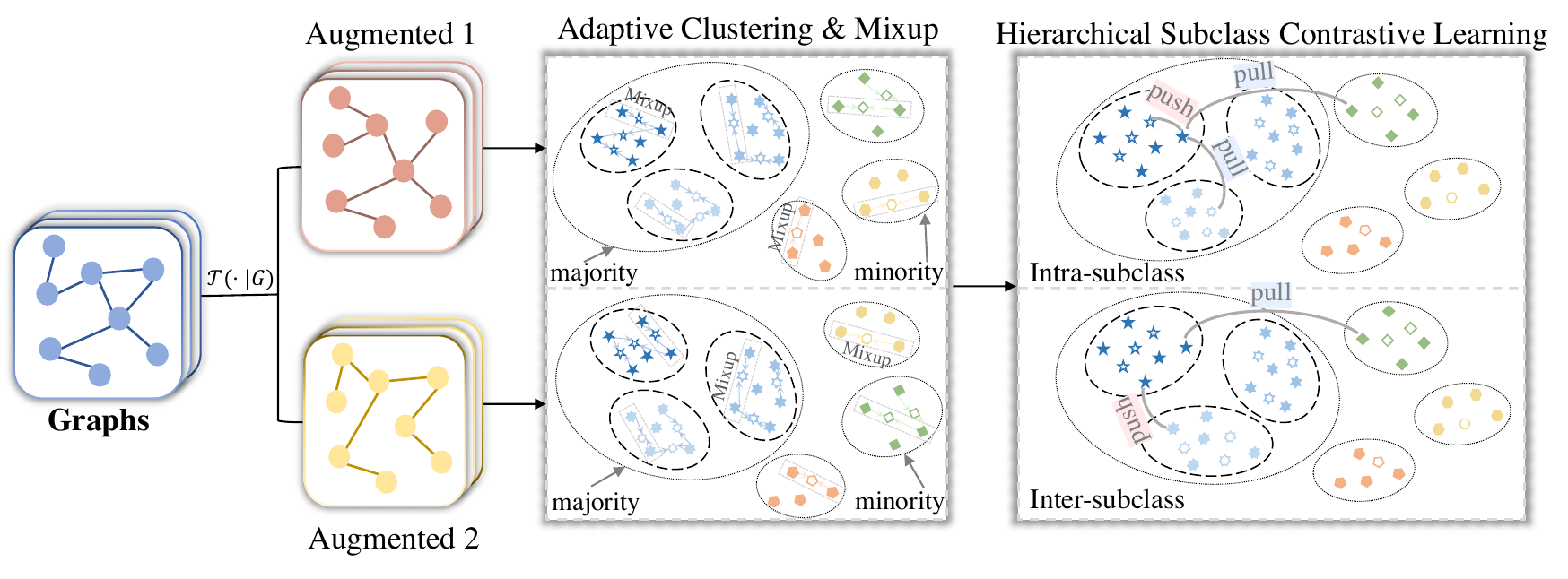}
    \caption{{Illustration of the proposed framework \method{}. 
    }}
    \label{fig:framework}
\end{figure*}


\noindent\textbf{Notations.} 
Given a graph dataset $\mathcal{G}=\{G_i, y_i\}_{i=1}^N$ comprising $N$ graphs, each associated with a ground-truth class label $y_i \in \{1,\ldots, K\}$. Without loss of generality, denote $N_j$ the number of graphs in the $j$-th class, and assume that $N_1\geq N_2 \geq \cdots \geq N_K$ following a descending order. To quantify the degree of class imbalance, we define the imbalance factor (IF) of the dataset as the ratio $N_1/N_K$. A class-imbalanced dataset consists of an imbalanced training set and a balanced test set, where the training set satisfies:
\begin{equation}
\label{LT}
\small
\begin{cases}
    \displaystyle{\int p(G|y=k_1) dG} \geq \int p(G|y=k_2) dG, & \forall ~ k_1 \leq k_2, \\
    \lim_{k \rightarrow \infty} \displaystyle{\int p(G|y=k)dG = 0}, &
\end{cases}
\end{equation}
which reveals a successive decay in class volumes as the class indexes ascend, and the probability ultimately converges to zero in the final few classes.

\smallskip
\noindent\textbf{Class-imbalanced Graph Classification.} 
The objective is to develop an unbiased classifier based on the graph dataset characterized by imbalanced class distributions. The trained classifier should learn robust and discriminative graph representations, ensuring that it remains capable of accurately classifying graphs across all classes, without being dominated by the abundant majority classes. Moreover, the trained classifier should exhibit strong generalization capabilities when tested on a balanced dataset.

\smallskip
\noindent\textbf{GNN-based Encoder.} Recent GNNs leverage both the graph structure and node features to learn an effective representation for a given graph via message-passing mechanism, which entails iteratively updating the embedding of a node $\mathbf{h}_v$ by aggregating embeddings from its neighboring nodes: 
\begin{equation}
\mathbf{h}_v^l=\operatorname{AGGREGATE}\left(\mathbf{h}_v^{l-1}, \mathbf{h}_u^{l-1} \mid u \in \mathcal{N}(v)\right) \text {, }
\end{equation}
where $\mathbf{h}_v^l$ is the embedding of node $v$ at the $l$-th layer of the GNN. $\mathcal{N}(v)$ is the neighbors of node $v$, $\operatorname{AGGREGATE}$ refers to an aggregation function. By iterating $L$ times, the graph-level representation $\mathbf{h}_G$ can be obtained by aggregating node representations $\mathbf{h}_v$ using the $\operatorname{READOUT}$ function:
\begin{equation}
    \label{eq:gnn}
    \mathbf{h}_G=\operatorname{READOUT}\left(\mathbf{h}_v^L \mid v \in V\right)
\end{equation}
where $V$ is the node set of graph $G$. Afterward, the derived graph-level representation $\mathbf{h}_G$ can be well used for downstream graph-level classification.

\section{Methodology}
\label{sec::model}


Our \method{} mainly includes three modules, i.e., adaptive clustering for subclass-balancing, subclass mixup interpolation, and hierarchical subclass contrastive learning. The framework overview of our proposed method is illustrated in Figure \ref{fig:framework}. In the subsequent sections, we provide a detailed explanation of the three components.

\subsection{Adaptive Clustering for Subclass-balancing}

Most existing class-imbalanced methods typically sacrifice the performance of the majority classes to improve the accuracy of the minority classes~\cite{mao2023rahnet,yi2023towards}. However, we argue that due to the complexity and diversity of the graph topology, the samples in the majority class often exhibit rich hierarchical substructures. It is crucial to sufficiently explore the hierarchical semantic information within them and we achieve this by decomposing each majority class into multiple semantically coherent subclasses, thereby balancing the various classes.

To this end, we leverage the idea of clustering to adaptively decompose the graph samples of each majority class into multiple subclasses~\cite{hou2023subclass}. Technically, considering a majority class $c$ and its corresponding graph dataset $\mathcal{G}_c \subseteq \mathcal{G}$, we first feed these graphs to the GNN-based encoder to obtain graph-level representations $\mathcal{Z}^c = \{{\mathbf{z}}_1^c, {\mathbf{z}}_2^c, \cdots, {\mathbf{z}}_{N_c}^c\}$ by Eq.~\eqref{eq:gnn}, where $N_c$ is the number of graphs in class $c$. Then we employ a selected clustering algorithm (\emph{e.g.}, \emph{k}-means) based on the graph representations, to partition $\mathcal{G}_c$ into multiple subclasses. To ensure a balanced distribution of samples among the subclasses, we set a threshold value $M$ as an upper limit of sample size within each subclass defined as:
\begin{equation}
    M=\max \left(n_K, \delta\right)
\end{equation}
where $n_K$ represents the number of the last minority classes sorted in descending order. The hyperparameter $\delta$ serves to regulate the minimum sample size within clusters, thereby preventing the formation of excessively small clusters. Notably, the cluster centers via the clustering algorithm can be progressively updated every several epochs during the learning process, such that we can adaptively decompose the optimal subclasses. Additionally, we only apply the clustering algorithm to majority classes that contain multiple instances, while keeping the minority classes unchanged. Therefore, this approach ensures that the number of samples in each resulting subclass is approximately equal to that of the minority class, thus alleviating class imbalance.

In this way, we can achieve subclass balance by decomposing the majority class into multiple semantically coherent subclasses, while capturing the abundant hierarchical semantic substructures within the majority class.

\subsection{Subclass Mixup Interpolation}
After eliminating class imbalance through adaptive clustering, the subclasses of the majority class roughly have a similar number of samples as the minority class. However, due to the sparsity of the minority class, directly optimizing on this basis can easily cause the learned representations to collapse into trivial solutions, leading to severe performance degradation. Existing class-imbalanced methods such as re-sampling and re-weighting strategies~\cite{chawla2002smote,cui2019class}, often employ oversampling or increasing the weight of the minority class to enhance the emphasis on them. Nevertheless, a problem arises in that minority classes typically have a relatively simple semantic structure, making it difficult to represent all the minority classes by replicating or increasing the weight of these samples. This makes the learned representations sensitive and poorly generalized when faced with samples from other minority classes.

To address this issue, we introduce the Mixup technique~\cite{zhang2017mixup} to synthesize new samples within subclasses, thereby increasing the diversity of samples and enriching their semantic structure, as well as avoiding the collapse of representation caused by sample sparsity. Specifically, given a subclass $s$ from class $c$, its feature representations is represented as $\mathcal{Z}_{s}^c = \{{\mathbf{z}}_1^c, {\mathbf{z}}_2^c, \cdots, {\mathbf{z}}_{N_{c_s}}^c\}$, where $N_{c_s}$ is the number of graphs in subclass $s$ from class $c$. Then we conduct interpolation between the feature representations ${\mathbf{z}}_i^c$ and ${\mathbf{z}}_j^c$ within each subclass, denoted as:
\begin{equation}
\tilde{{\mathbf{z}}}^c=\alpha \cdot {\mathbf{z}}_i^c +(1-\alpha) \cdot {\mathbf{z}}_j^c, 
\end{equation}
whose subclass label is annotated to be the same as ${\mathbf{z}}_i^c$ and ${\mathbf{z}}_j^c$, and $\alpha$ is a scalar mixing ratio, sampled from a uniform distribution $U[0,1]$~\cite{zhang2017mixup}.

In this way, Mixup extends the training samples within each subclass by incorporating the prior knowledge based on the smoothing assumption, enhancing the diversity and semantic richness of samples in each subclass.

\subsection{Hierarchical Subclass Contrastive Learning}

By adaptively clustering the majority classes into multiple subclasses with similar sizes, we can now treat all subclasses (including the original minority classes) in a balanced manner. This allows us to address class imbalance while also harvesting abundant semantic information with different hierarchical structures: coarse-grained class labels and fine-grained subclass labels.

As such, inspired by the powerful representation learning capability of contrastive learning, which learns to discriminate the samples by contrasting the negative ones~\cite{chen2020simple,you2020graph,ju2024towards}. We leverage supervised contrastive learning \cite{khosla2020supervised} to effectively learn discriminative representations from both \emph{intra-subclass} and \emph{inter-subclass} perspectives.

From the \emph{intra-subclass} view, we aim to encourage a graph to be closer to graphs from the same subclass than to any other graph. Specifically, given a batch  $\mathcal{B}=\{G_i, \Omega(G_i)\}_{i=1}^B$ where $\Omega(G)$ denotes the subclass label of graph $G$. To conduct effective graph contrastive learning, for each graph $G$ we first perform stochastic graph augmentations $\mathcal{T}(\cdot|G)$ to obtain two correlated views $\hat{G}^1$ and $\hat{G}^2$ as a positive pair, where $\mathcal{T}(\cdot|G)$ is pre-defined augmentation distribution (\emph{e.g.}, \emph{node dropping}, \emph{edge perturbation}, \emph{attribute masking}, \emph{subgraph} following \cite{you2020graph}). Then, graph-level representations of the resulting $2B$ augmented graphs are extracted by the GNN-based encoder denoted as $\{\mathbf{z}_1, \mathbf{z}_2, ...,\mathbf{z}_{2B}\}$ with an abuse of notation. Let $i\in I=\{1,...,2B\}$ be the index of an arbitrary augmented graph in a batch, we formulate the supervised contrastive loss for $i$-th graph from the \emph{intra-subclass} view as:
\begin{equation}
    \label{eq:intra}
    \mathcal{L}_i^{\text{intra}}= -\frac{1}{| Q(i)|} \sum_{j \in Q(i)} \log \frac{e^ {\mathbf{z}_i^{\top} \mathbf{z}_j / \tau}}{\sum\limits_{a \in A(i)} e^{\mathbf{z}_i^{\top} \mathbf{z}_a / \tau}}
\end{equation}  
where $A(i)= I\setminus\{i\}$, $Q(i)=\{q\in A(i):\Omega(G_q)=\Omega(G_i)\}$ is the set of indices of all positives distinct from $i$ with the same subclass label as $G_i$, and $\tau$ is a scalar temperature hyper-parameter.

From the \emph{inter-subclass} view, we aim to encourage a graph to be closer to graphs from different subclasses but the same class than to graphs from any other subclass. Analogously, we formulate the supervised contrastive loss for $i$-th graph from the \emph{inter-subclass} view as:
\begin{equation}
    \label{eq:inter}
    \mathcal{L}_i^{\text{inter}}= -\frac{1}{|P(i)|-|Q(i)|} \sum_{j \in P(i)/Q(i)} \log \frac{e^ {\mathbf{z}_i^{\top} \mathbf{z}_j / \tau}}{\sum\limits_{a \in A(i)/Q(i)} e^{\mathbf{z}_i^{\top} \mathbf{z}_a / \tau}}
\end{equation}  
where $P(i)=\{p\in A(i):y_q=y_i\}$ is the set of indices of all positives distinct from $i$ with the same class label as $G_i$.

Notably, Eq.~\eqref{eq:intra} achieves subclass-level balance, where positive graphs are pairs within the same subclass, enabling the model to understand the intra-subclass relationships. On the other hand, Eq.~\eqref{eq:inter} achieves class-level balance, which leverages the class information but excludes graphs from the same subclass to focus on those within the same class but belonging to different subclasses, facilitating the learning of inter-subclass relationships from complementary view.

\smallskip
\noindent\textbf{Joint Optimization.}
To enhance class-imbalanced graph classification by learning effective representations in a hierarchical way, we formally combine the two supervised contrastive losses from both \emph{intra-subclass} and \emph{inter-subclass} perspectives in a batch, which is defined as:
\begin{equation}
    \label{eq:total_L}
    \mathcal{L}=\sum_{i=1}^B (\mathcal{L}_i^{\text{intra}} + \beta \cdot \mathcal{L}_i^{\text{inter}}), 
\end{equation}
where $\beta$ is the balance hyper-parameter to adjust the relative importance of each loss component, and we set it to 1 by default in the experiments. We summarize the whole optimization algorithm of \method{} in Algorithm \ref{alg:algorithm}.

\begin{algorithm}[tb]
    \caption{Optimization Algorithm of \method{}}
    \label{alg:algorithm}
    \textbf{Input}: Class-imbalanced graph dataset $\mathcal{G}=\{G_i, y_i\}_{i=1}^N$, time interval of updating cluster centers $T$, control parameter $\delta$, and temperature parameter $\tau$\\
    \textbf{Output}: Balanced classifier

    \begin{algorithmic}[1] 
    \STATE Initialize GNN-based encoder parameter.\\
    \STATE Train GNN in the first few epochs for warm-up.\\
    \WHILE{\emph{not done}}
    \STATE Adaptively update the cluster centers based on the current graph representations every $T$ epochs.
    \STATE Sample one augmentation from \citet{you2020graph}.
    \STATE Compute supervised contrastive loss $\mathcal{L}_i^{\text{intra}}$ for intra-subclass by Eq.~\eqref{eq:intra}.
    \STATE Compute supervised contrastive loss $\mathcal{L}_i^{\text{inter}}$ for inter-subclass by Eq.~\eqref{eq:inter}.
    \STATE Update GNN parameter by gradient descent to minimize $\mathcal{L}$ by Eq.~\eqref{eq:total_L}.
    \ENDWHILE
    \end{algorithmic}
\end{algorithm}

\subsection{Computational Complexity Analysis} 

With $B$ as the batch size and $\left| V \right|$ as the average number of nodes in input graphs, obtaining embeddings from the GNN encoder has a time complexity of $O(N D \left| V \right|)$, where $N$ is the number of graphs, and $D$ is the embedding dimension. The time complexity of subclass clustering is $O(I K N D)$, where $I$ is the iterations until convergence and $K$ is the total number of clusters. Since clustering algorithms converge quickly in practice, $I$ and $K$ can be considered constants, and the complexity can be simplified to $O(ND)$. Subclass contrastive learning computes the loss in $O(N D B)$ time. Therefore, the overall time complexity of \method{} is $O(N D (\left| V \right| + B))$.


\section{Experiment}
\label{sec::experiment}

\begin{table*}[t]
\begin{center}
\caption{Overall performance ($\%$) with various IFs on six benchmark datasets for class-imbalanced graph classification. The best results are shown in boldface and the second-best results are underlined.}
\label{table:experiment}
\resizebox{0.98\textwidth}{!}{ %
\begin{tabular}{lcccccccccccc}
\toprule
Model &\multicolumn{2}{c}{Synthie} &\multicolumn{2}{c}{ENZYMES}  &\multicolumn{2}{c}{MNIST} &\multicolumn{2}{c}{Letter-high} &\multicolumn{2}{c}{Letter-low} & \multicolumn{2}{c}{COIL-DEL}\\
& IF=15 & IF=30 & IF=15 & IF=30  & IF=50 & IF=100 & IF=25 & IF=50  & IF=25 & IF=50 & IF=10 & IF=20\\ 
\midrule
GraphSAGE      &34.74  &30.25  &30.66  &25.16  &68.67  &63.46  &51.06  &42.16  &86.00  &84.32  &38.80  &31.32 \\
Up-sampling      &35.25  &33.50  &32.33  &28.50  &64.69  &59.78  &53.62  &44.20  &88.48  &86.72  &39.20  &26.96 \\
\midrule
CB loss         &34.75  &30.75  &32.19  &26.83  &68.85  &63.40  &53.76  &45.06  &87.46  &85.44  &41.72  &32.34 \\
LACE loss       &33.25  &30.85  &31.16  &25.50  &69.72  &64.59  &47.46  &38.94  &87.89  &84.69  &41.96  &32.18 \\
\midrule
GraphCL     &40.25   &36.25   &36.66  &29.83    &69.37  &65.12  &57.34  &48.93  &89.28  &87.89  &42.02   &33.19 \\
SupCon      &40.34   &37.25   &37.08  &30.67    &69.76  &64.88  &57.29  &48.93  &89.12  &87.36  &42.93   &34.20 \\
\midrule
Augmentation    &39.37  &35.37  &32.08  &26.75  &72.18  &68.17  &49.28  &42.36  &88.32  &86.40  &38.18  &30.80 \\
G$^2$GNN$_{n}$     &38.08  &27.94  &35.00  &29.17  &70.91  &66.73  &58.91  &\underline{51.12}  &89.49  &87.98  &38.32  &27.98 \\
G$^2$GNN$_{e}$     &40.19  &\underline{37.53}  &35.83  &29.50  &73.69  &70.31  &58.85  &49.96  &\underline{89.84}  &87.80  &39.18  &31.06 \\ 
RAHNet  &\underline{42.35}  &36.76  &\underline{38.50}  &\underline{32.17}  &\underline{75.12}  &\underline{71.98}  &\underline{59.20}  &50.37  &89.65  &\underline{88.69}  &\underline{43.04}  &\underline{36.80} \\
\midrule
\method{} (Ours) &\textbf{43.60} &\textbf{39.69} &\textbf{39.01} &\textbf{32.41} &\textbf{77.38} &\textbf{73.62} &\textbf{59.88} &\textbf{54.38} &\textbf{92.08} &\textbf{90.52} &\textbf{47.54} &\textbf{38.23}\\
\bottomrule
\end{tabular}
} 
\end{center}
\end{table*}

\subsection{Experimental Settings}

\noindent\textbf{Datasets.}
We evaluate the effectiveness of our proposed model by examining it on both synthetic and real-world datasets from various domains. Specifically, the datasets are categorized into three groups: (a) synthetic: Synthie \cite{morris2016faster}, (b) bioinformatics: ENZYMES \cite{schomburg2004brenda}, and (c) computer vision: MNIST \cite{dwivedi2020benchmarking},  Letter-high \cite{riesen2008iam}, Letter-low \cite{riesen2008iam}, and COIL-DEL \cite{riesen2008iam}. To ensure that the datasets follow Zipf's law exactly \cite{newman2005power}, the training sets were transformed into class-imbalanced datasets with varying imbalance factors (IFs), while the validation and test sets remained balanced. The dataset is split into training, validation, and testing sets with a 6:2:2 proportion for each respective set.

\noindent\textbf{Baselines.}
We compare the effectiveness of our framework \method{} against several competitive class-imbalanced graph classification baselines. These baselines can be categorized into four perspectives: (a) Data re-sampling methods: up-sampling \cite{chawla2003c4}; (b) Loss re-balancing methods: class-balanced (CB) loss \cite{cui2019class} and logit adjusted cross-entropy (LACE) loss \cite{menon2020long}; (c) Contrastive learning based methods: graph contrastive learning (GraphCL) \cite{you2020graph} and supervised contrastive learning (SupCon) \cite{khosla2020supervised}; (d) Information augmentation methods: graph augmentation \cite{yu2022graph}, G$^2$GNN \cite{wang2022gog} and RAHNet~\cite{mao2023rahnet}.

\noindent\textbf{Implementation details.}
In our experiments, we utilized GraphSAGE \cite{hamilton2017inductive} as the GNN backbone encoder with a two-layer MLP classifier. The models were optimized using the Adam optimizer with a fixed learning rate of 0.0001 and a batch size of 32. For our \method{}, we set the temperature parameter $\tau$ to 0.2 and set the time interval $T$ of dynamic updating the cluster centers to 10. Moreover, we fine-tuned the cluster size control parameter $\delta$ for each dataset individually. We evaluated the models based on the average top-1 accuracy over 10 run times.

\subsection{Experimental results}

We record the class-imbalanced classification accuracy of our \method{} and the aforementioned baseline methods on six graph classification benchmarks w.r.t. different imbalanced factors.
The results in Table~\ref{table:experiment} reveals the following insights:

(i) The performance of both baselines and our \method{} significantly decreases as the imbalance between majority and minority classes increases.
This finding supports our motivation that GNNs struggle in imbalanced settings and are especially vulnerable to imbalanced class distributions.

(ii) Information augmentation methods, such as two G$^2$GNN variants and RAHNet, demonstrate superior performance compared to data re-sampling and loss re-balancing methods. This suggests that information augmentation methods introduce new information to improve the representation of minority classes.
Furthermore, RAHNet exhibits stable performance across most datasets, as it separately trains a feature extractor with graph retrieval for intra-class diversity and a classifier with Max-norm and weight decay for balanced weights in long-tailed scenarios. 

(iii) The table reveals that our \method{} outperforms all baseline methods on all six datasets. 
This remarkable performance can be primarily attributed to that \method{} adaptively clusters the majority classes into multiple subclasses of similar sizes, which enables balanced treatment of all subclasses during model training and effectively addresses the issue of class imbalance in graphs. Additionally, the introduction of the novel subclass mixup interpolation further enhances sample diversity and enriches the semantic structure by synthesizing new samples within the subclasses.

\subsection{Ablation study}

\begin{table}[t]
\LARGE
\setlength\tabcolsep{2pt}  
\begin{center}
\caption{Ablation study results of several model variants (Hierarchical Subclass Contrastive Learning: SCL; Adaptive Clustering: AC; Subclass Mixup Interpolation: SMI).}
\label{table:ab}
\resizebox{1.0\linewidth}{!}{
     \begin{tabular}{lcccccc}
        \toprule
        Methods &Synthie &ENZYMES &MNIST &Letter-high &Letter-low &COIL-DEL \\
        \midrule
        \method{} w/o HSCL &40.57 &36.92 &74.23 &56.40 &90.10 &45.82 \\
        \method{} w/o AC &42.33 &38.84 &76.20 &58.91 &91.24 &46.40 \\
        \method{} w/o SMI &43.08 &38.43 &77.25 &57.73 &91.16 &46.28 \\
        \midrule
        Complete Model &43.60 &39.01 &77.38 &59.88 &92.08 &47.54 \\
        \bottomrule
    \end{tabular}
}
\end{center}
\end{table}

In this part, we conducted comprehensive ablation studies on all six datasets to demonstrate the effectiveness of the proposed \method{}. We investigated three variants of \method{}: (i) \method{} w/o HSCL, which removes the hierarchical subclass contrastive module and only utilizes supervised loss; (ii) \method{} w/o AC, which excludes the adaptive updating of cluster centers; (iii) \method{} w/o SMI, which removes the data enrichment from mixup interpolation.

Based on the results in Table \ref{table:ab}, it is evident that the model experiences a substantial decline in performance when the subclass contrastive learning module is removed. This indicates that our subclass contrastive module balances all subclasses, allowing the model to effectively learn discriminative representations from both intra- and inter-subclass perspectives. We also notice that the absence of adaptive clustering leads to performance decreases in all cases. This can be attributed to adaptive clustering dynamically adjusting cluster assignments based on the current learned model, which tends to improve as the training progresses. Furthermore, the use of mixup interpolation proves to be beneficial for overall performance by effectively enriching sample diversity and mitigating the issue of data scarcity.





\subsection{Hyper-parameter Sensitivity}
\label{sec::hyper}

\begin{figure}[t]
\centering
\resizebox{0.99\linewidth}{!}{
    \subfloat[Synthie]{
     \includegraphics[width = 0.33\linewidth]{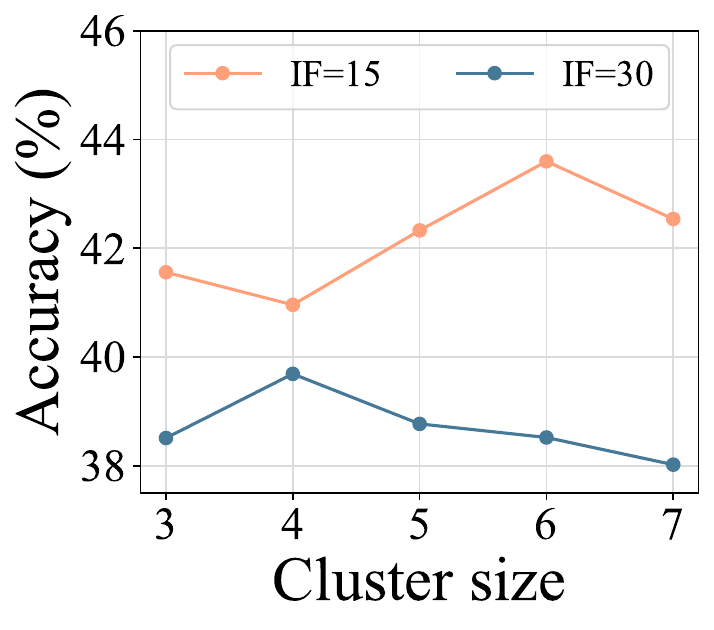}}
     \hfill
     \subfloat[ENZYMES]{
     \includegraphics[width = 0.33\linewidth]{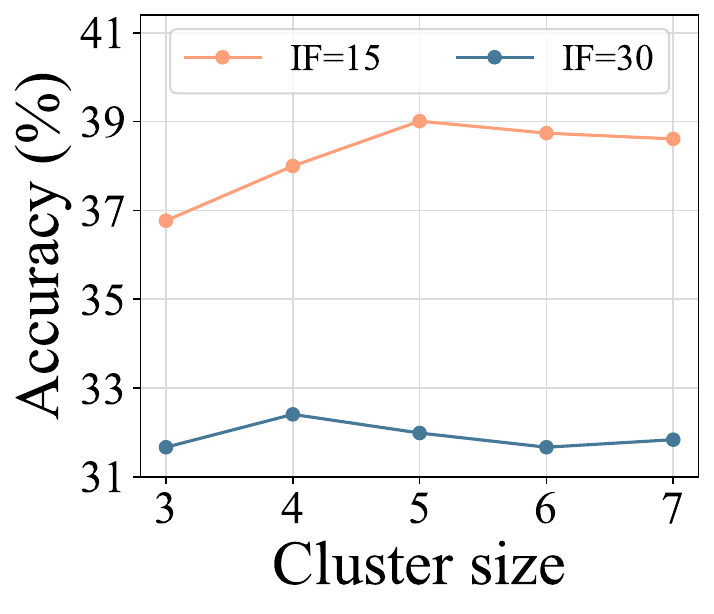}}
     \hfill
     \subfloat[MNIST]{
     \includegraphics[width = 0.33\linewidth]{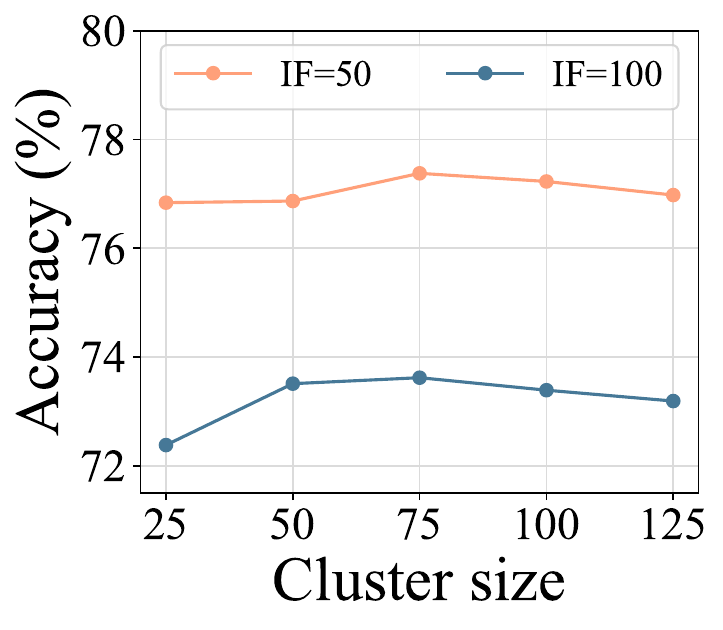}}
    }

\resizebox{0.99\linewidth}{!}{
     \subfloat[Letter-high]{
     \includegraphics[width = 0.33\linewidth]{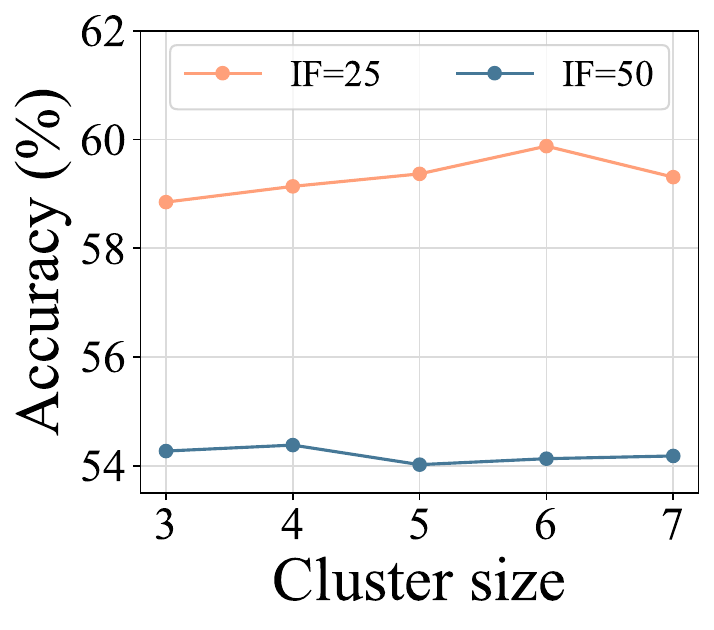}}
     \hfill
     \subfloat[Letter-low]{
     \includegraphics[width = 0.33\linewidth]{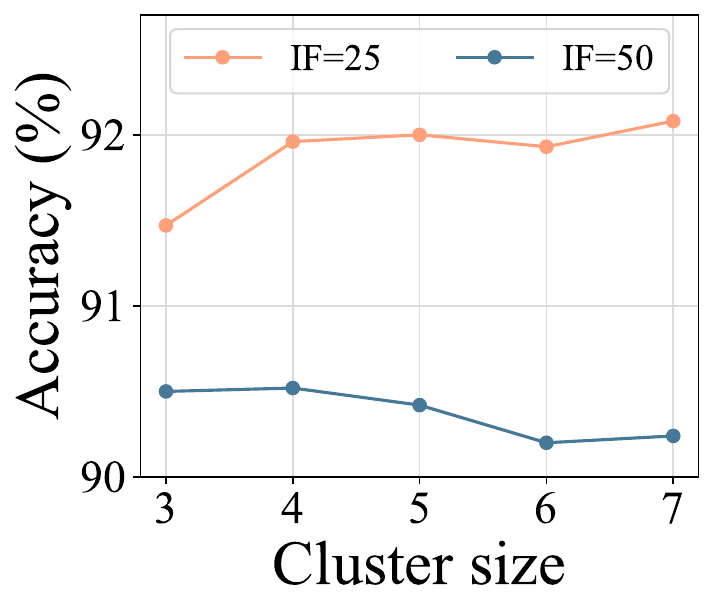}}
     \hfill
     \subfloat[COIL-DEL]{
     \includegraphics[width = 0.33\linewidth]{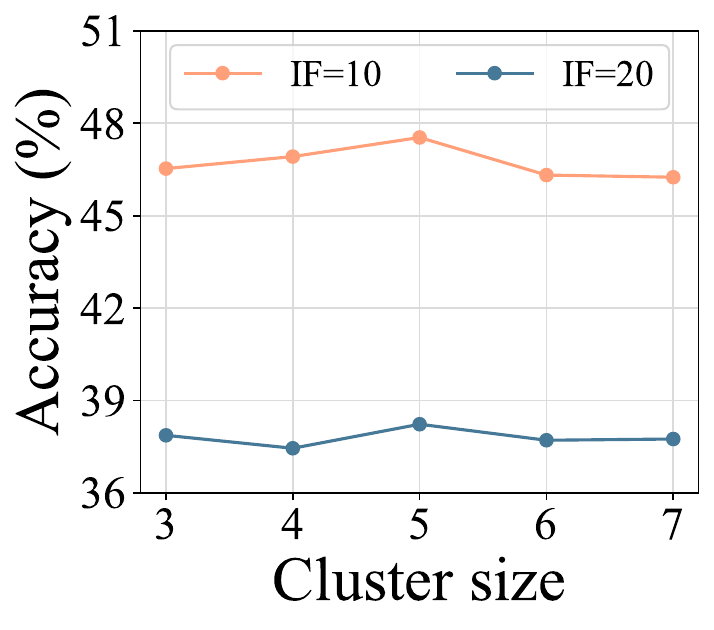}}
    }
\caption{Sensitivity analysis of cluster size.}
\label{fig:params}
\end{figure}

Here we investigate the sensitivity of our proposed \method{} to hyper-parameters. Specifically, we study the influence of varying different cluster sizes on six datasets, which are essential to contrastive learning performance. We vary the cluster size in the range of $\{$25, 50, 75, 100, 125$\}$ for the MNIST dataset, and in the range of $\{$3, 4, 5, 6, 7$\}$ for other datasets. The results are shown in Figure \ref{fig:params}. 

The accuracy curves exhibit a similar trend across all datasets. We can observe that the performance of our \method{} gradually improves as the cluster size increases until it reaches saturation. However, choosing a cluster size that is too large or too small may have a detrimental effect on model performance. We argue that this is because a large cluster size can result in some subclasses containing significantly more instances than the minority classes, leading to an imbalance among subclasses, which in turn leads to suboptimal performance. Moreover, a small cluster size can result in similar instances being assigned to different clusters, which can adversely affect the learned representations and result in a performance drop. In addition, it can be noticed that datasets with smaller IFs tend to have larger optimal cluster sizes compared to those with larger IFs.  For instance, the Letter-high dataset achieves the best performance when the cluster size is set to 6 for an IF of 25, whereas the optimal cluster size reduces to 4 as the IF increases to 50. This suggests that the ideal cluster size is influenced by the number of instances in the minority class.

\subsection{Analysis of Feature Distribution}


\begin{figure}[t]
\centering
\resizebox{0.99\linewidth}{!}{
     \subfloat[Feature distance in classes]{
     \label{fig::classes}
     \includegraphics[width = 0.5\linewidth]{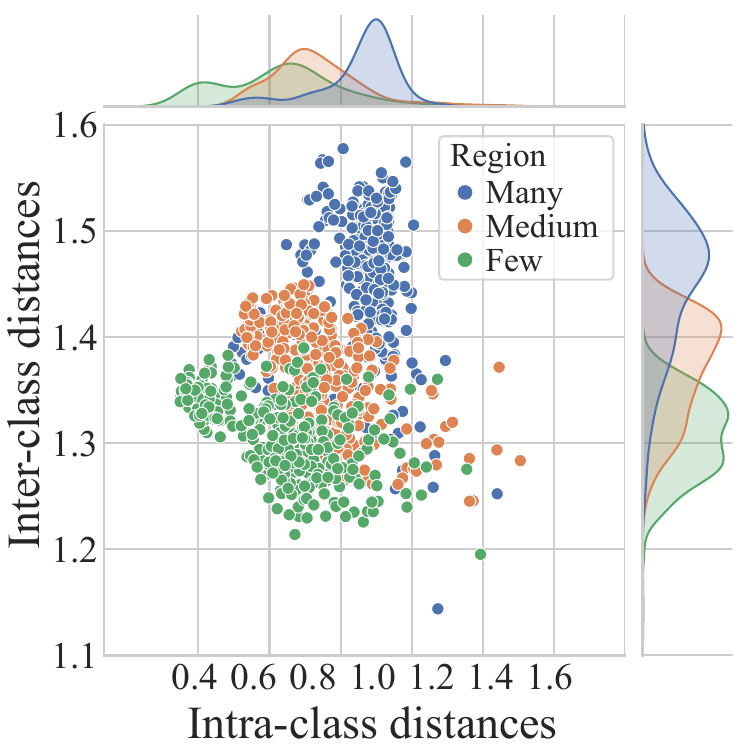}}
     \hfill
     \subfloat[Feature distance in subclasses]{
     \label{fig::subclasses}
     \includegraphics[width = 0.5\linewidth]{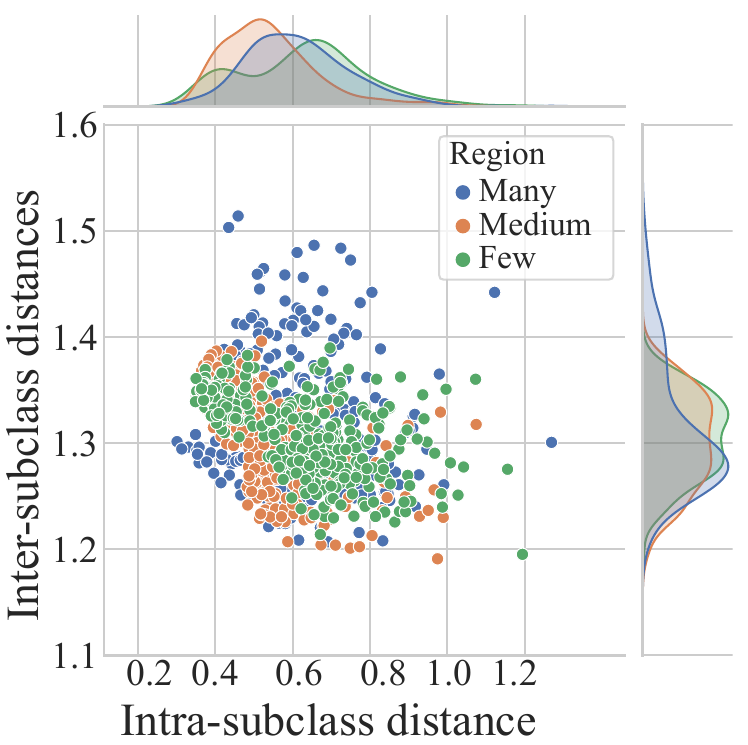}}
    }
\caption{Feature distance and distribution curve of classes and subclasses on MNIST with imbalance factor 50.}
\label{fig:distance}
\end{figure}

To analyze the learned representation of \method{}, we begin by defining the intra-class and inter-class distances using the Euclidean distance between a given sample and other samples from the same or different classes. Specifically, we calculate the Euclidean distance between a sample $z_i$ and a set $S$ as $\boldsymbol{D}(z_i, S) = \frac{1}{\left| S \right|} \sum_{z_j \in S} \Vert z_i - z_j \Vert_2$. Next, we define the intra- and inter-class distance of sample $z_i$ as $\boldsymbol{D}(z_i, P(i))$ and $\boldsymbol{D}(z_i, \mathcal{G}/P(i))$, respectively. Similarly, we define the intra-and inter-subclass distance of sample $z_i$ as $\boldsymbol{D}(z_i, Q(i))$ and $\boldsymbol{D}(z_i, P(i)/Q(i))$, respectively.


We visualize the feature distance distributions using the MNIST dataset with an imbalance factor of 50. We first divided the classes into the many, medium, and few regions based on the number of samples. Figure \ref{fig::classes} demonstrates the intra-class/inter-class feature distances among different data groups and their corresponding distributions, while Figure \ref{fig::subclasses} displays the intra-subclass/inter-subclass feature distances. From the results, we can draw several conclusions: (i) The average intra-subclass distance is lower than the average intra-class distance, which implies that \method{} encourages samples from the same subclass to have similar representations. This observation suggests that \method{} successfully captures and emphasizes the finer distinctions present within each class. (ii) The feature distances in subclasses exhibit a higher degree of uniformity across different data subsets compared to the feature distances in classes. This key finding indicates that the subclasses belonging to both majority and minority classes occupy similar volumes of the learned feature space. Consequently, this balance among the subclasses contributes to the overall class balance.

\subsection{Visualization of Clustering Results}

\begin{figure}[t]
\centering
\resizebox{1.0\linewidth}{!}{
     \subfloat[Subclass:3A]{
     \includegraphics[width = 0.25\linewidth]{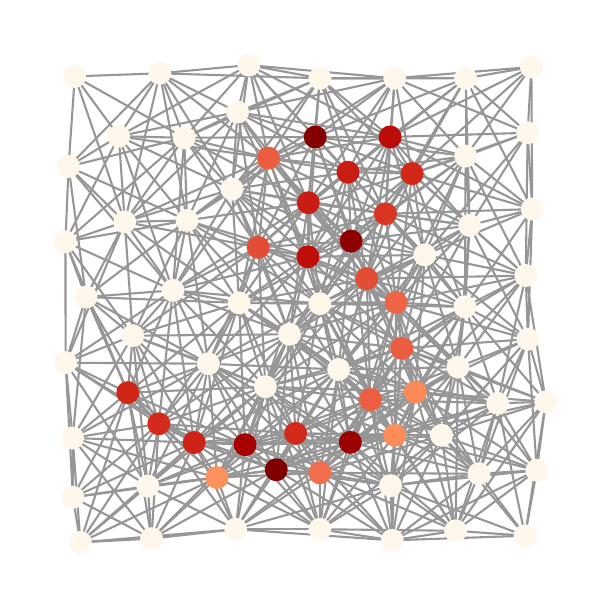}}
     \hfill
     \subfloat[Subclass:3A]{
     \includegraphics[width = 0.25\linewidth]{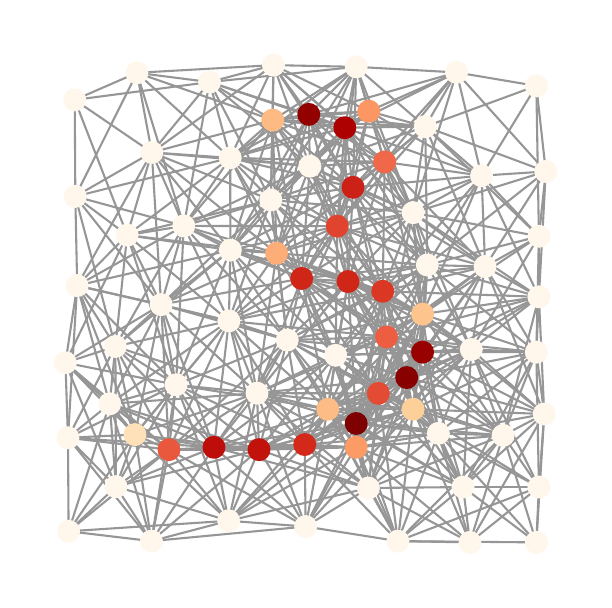}}
     \hfill
     \subfloat[Subclass:3B]{
     \includegraphics[width = 0.25\linewidth]{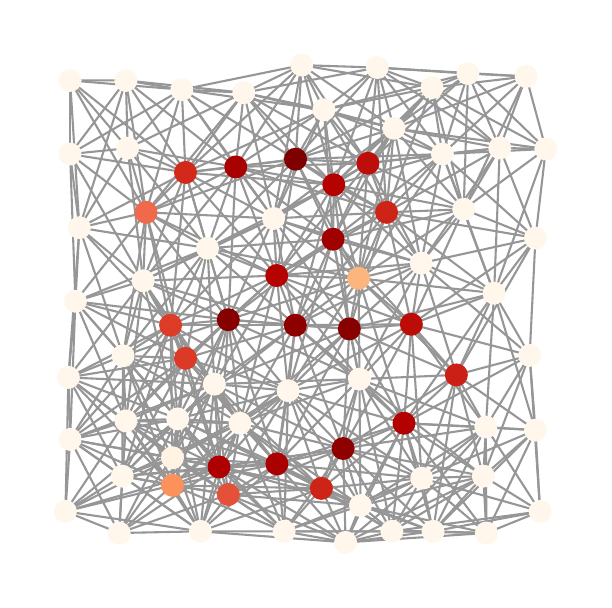}}
     \hfill
     \subfloat[Subclass:3B]{
     \includegraphics[width = 0.25\linewidth]{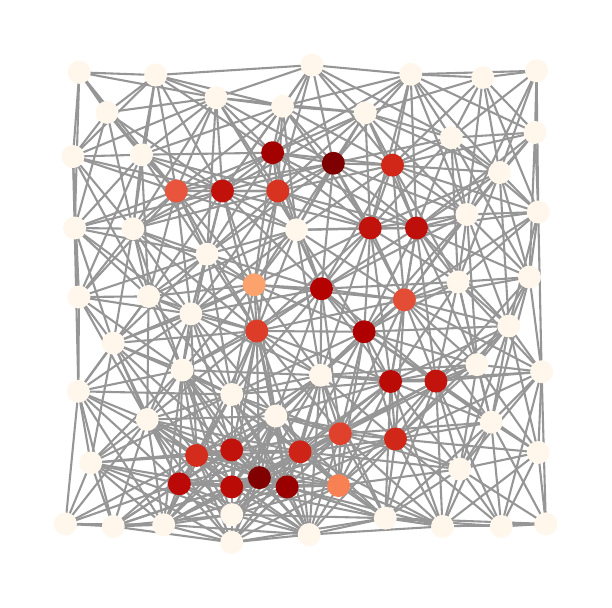}}
    }

\resizebox{1.0\linewidth}{!}{
     \subfloat[Subclass:7A]{
     \includegraphics[width = 0.25\linewidth]{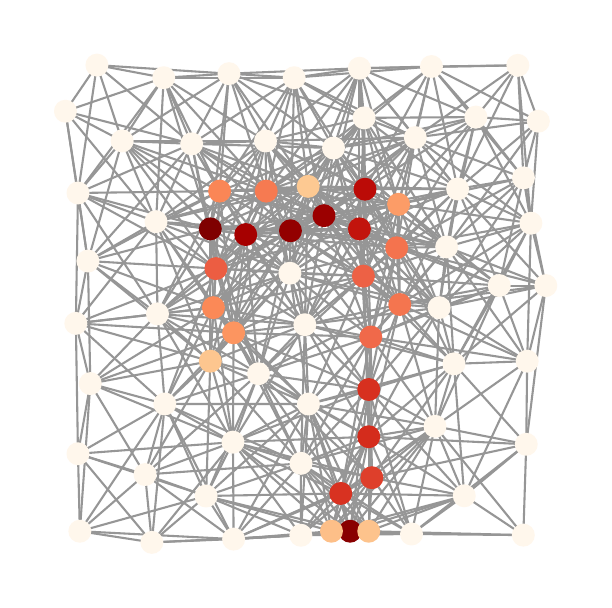}}
     \hfill
     \subfloat[Subclass:7A]{
     \includegraphics[width = 0.25\linewidth]{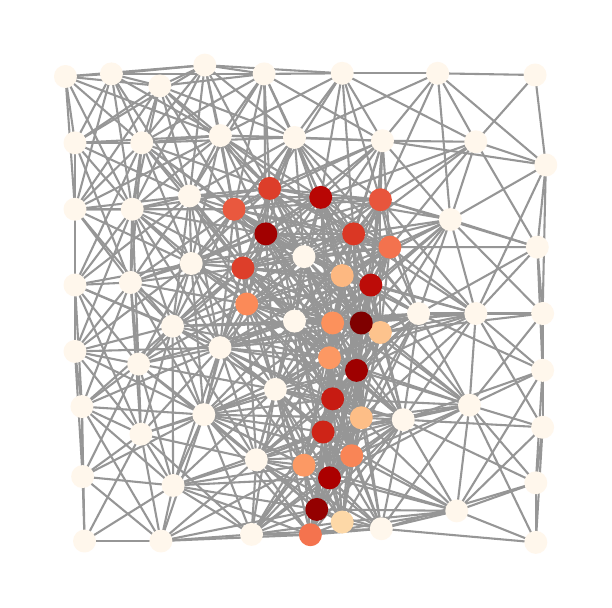}}
     \hfill
     \subfloat[Subclass:7B]{
     \includegraphics[width = 0.25\linewidth]{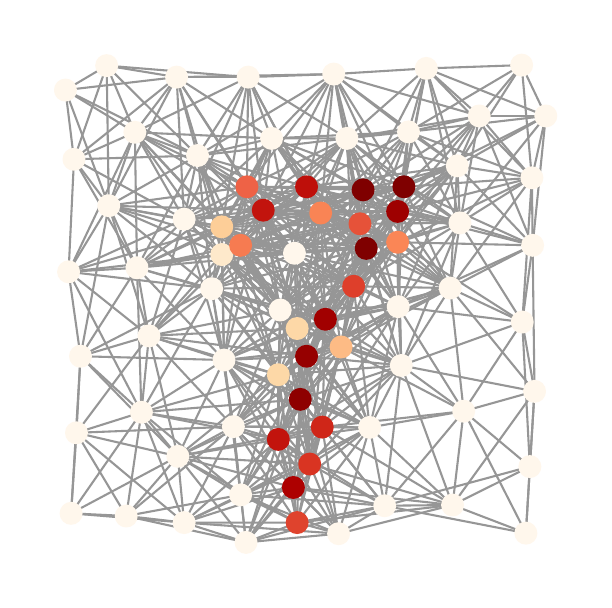}}
     \hfill
     \subfloat[Subclass:7B]{
     \includegraphics[width = 0.25\linewidth]{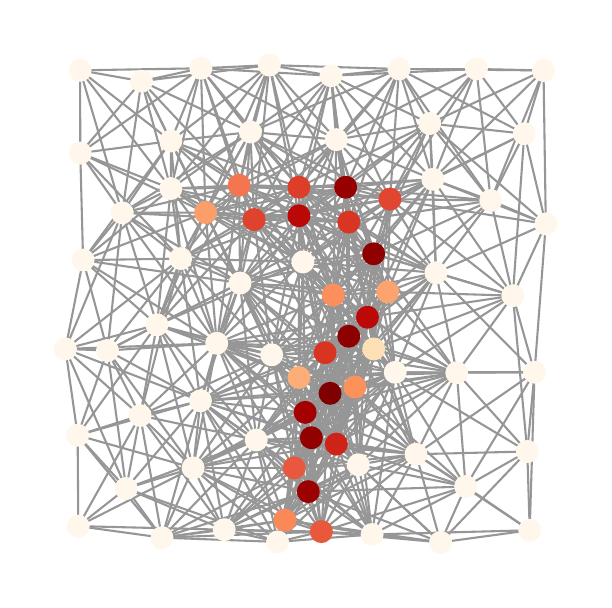}}
    }
\caption{Visualization of clustering results. We visualize several graphs in the MNIST dataset with their corresponding ground truth classes and resulting subclasses. Note that graphs (a) to (d) belong to class 3, where (a) and (b) belong to a different subclass from (c) and (d). On the other hand, graphs (e) to (h) are grouped under class 7, where (e) and (f) belong to a different subclass from (g) and (h).}
\label{fig:visualization}
\end{figure}

In this subsection, we visualize the clustering results to demonstrate the effectiveness of our subclass clustering and subclass contrastive module. Figure \ref{fig:visualization} illustrates eight graphs drawn from the MNIST dataset, where the nodes represent super-pixels of images and the colors indicate the grey-scale values of the super-pixels. Notably, while graphs within the same class may not exhibit significant structural similarities, those within the same subclass demonstrate remarkable resemblances. For instance, graph (e) and graph (f) show distinct writing styles compared to graph (g) and graph (h), thereby representing different variations within class 7. The clustering results emphasize the effectiveness of our proposed subclass clustering module, which successfully identifies and groups graphs with similar topological structures within specific subclasses. Moreover, our subclass contrastive module not only mitigates the negative impact of class imbalance while preserving sample balance but also enables the representations to cluster into finer-grained subclasses within the representation space. This demonstrates the capability of our approach to capture and leverage the subtle variations and distinctions among different subclasses, leading to more informative representations and enhanced classification performance.
\section{Related Work}
\label{sec::related}


\noindent\textbf{Graph Classification}
poses a pivotal challenge within the domain of graph analysis, aiming to recognize the class label for an entire graph. Existing methodologies for graph classification can be broadly classified into two fundamental streams: graph kernel-based approaches and GNN-based techniques. The former aims to measure the similarity between graphs by decomposing them into substructures (e.g., shortest paths~\cite{kashima2003marginalized}, graphlets~\cite{shervashidze2009efficient}, or subtrees~\cite{shervashidze2011weisfeiler}). The latter has gained significant attention in recent years due to the great capability to capture both structural and attribute information~\cite{ju2022tgnn,ju2024hypergraph,luo2024gala,luo2023towards}, whose key idea is to iteratively update node features by aggregating information from neighboring nodes~\cite{gilmer2017neural}. Despite the tremendous success, these methods are prone to prediction bias for imbalanced data while our \method{} alleviates this issue by clustering to balance the quantities of different classes.

\smallskip
\noindent\textbf{Class-imbalanced Learning}
(also known as long-tailed learning) endeavors to mitigate the influence of imbalanced class distributions, and there are broadly three main strategies: re-sampling~\cite{chawla2002smote,han2005borderline,guo2021long}, re-weighting~\cite{cui2019class,cao2019learning,hou2023subclass}, and ensemble learning~\cite{xiang2020learning,wang2020long,yi2023towards,mao2024learning}. To facilitate class-imbalanced learning for graph-structured data, many GNN-based methods have been proposed for node classification~\cite{shi2020multi,liu2021tail,park2021graphens,song2022tam,yun2022lte4g,zhou2023graphsr,zeng2023imgcl,ju2024survey}. However, these methods often overlook capturing the rich semantic structure of the majority classes and excessively focus on learning from the minority classes. Furthermore, existing GNN methods are designed for node classification, while our approach \method{} explores the unexplored domain of class-imbalanced graph classification.

\smallskip
\noindent\textbf{Contrastive Learning}
have garnered considerable attention due to their remarkable performance in representation learning and downstream tasks~\cite{chen2020simple,he2020momentum,khosla2020supervised,luo2024semievol}. The widely adopted loss function in this domain is the InfoNCE loss~\cite{oord2018representation}, which effectively pulls together augmentations of the same samples while simultaneously pushing away negative samples. Moreover, a multitude of recent advancements have emerged that extend contrastive learning to the graph domains~\cite{you2020graph,chen2023attribute,gong2023ma,gu2024deer,ju2024towards}. Different from the above works, our \method{} leverages supervised contrastive learning to hierarchically learn graph representations, thereby mitigating class imbalance.

\section{Conclusion}
\label{sec::conclusion}

In this paper, we focus on class-imbalanced graph classification and propose a novel framework \method{} to fully capture the semantic substructures within the majority class while effectively mitigating excessive focus towards the minority class. We first leverage clustering to balance the imbalance among different classes. Then we employ mixup to alleviate data sparsity and enrich intra-subclass semantics. Finally, we utilize supervised contrastive learning to effectively learn hierarchical graph representations from both intra-subclass and inter-subclass views. Extensive experiments demonstrate the superiority of \method{} over the baseline methods in a series of real-world graph benchmarks.

\section{Acknowledgments}
This paper is partially supported by the National Natural Science Foundation of China (NSFC Grant Numbers 62306014 and 62276002), the Postdoctoral Fellowship Program (Grade A) of CPSF with Grant No. BX20240239, the China Postdoctoral Science Foundation with Grant No. 2024M762201 as well as Sichuan University Interdisciplinary Innovation Fund.
\bibliography{aaai25}

\end{document}